\batchmode
\documentclass{article}

\usepackage{fancyvrb, natbib}
\usepackage[dvips]{graphicx}

\begin{document}

\title{Unsupervised Learning in a Framework of Information
Compression by Multiple Alignment, Unification and Search} 

\author{J Gerard Wolff \\
\\
CognitionResearch.org.uk,\\
Email: gerry@cognitionresearch.org.uk.}

\date{11 February 2003}

\maketitle

\begin{abstract}

This paper describes a novel approach to unsupervised learning that has been developed within a framework designed to integrate learning with such things as parsing and production of language, fuzzy pattern recognition and best-match information retrieval, class hierarchies with inheritance of attributes, probabilistic and exact forms of reasoning, and others. This framework, called {\em information compression by multiple alignment, unification and search} (ICMAUS), is founded on principles of Minimum Length Encoding pioneered by Solomonoff and others. This work originated in an earlier programme of research into unsupervised learning of natural languages and many of the goals remain the same: to use a framework for knowledge representation that is powerful enough to accommodate the structure of natural languages, to integrate the learning of segmental structure with the learning of disjunctive (class) structures, to distinguish `correct' generalisations from `over' generalisations without negative samples or external error correction or the grading of samples from simple to complex, learning `correct' forms despite corrupted data, and others. The main body of the paper describes SP70, a computer model of the ICMAUS framework that incorporates processes for unsupervised learning. Examples are presented to show how the model can infer plausible grammars from appropriate input. Limitation of the current model and how they may be overcome are briefly discussed.
\\
\\
\noindent {\em keywords}: unsupervised learning, information compression, multiple alignment, unification, minimum length encoding.

\end{abstract}

\section{Introduction}

This paper describes a novel approach to unsupervised learning that has been developed within 
a research programme whose overarching goal is the {\em integration} of diverse functions---learning, recognition, reasoning and others---within one relatively simple framework. This has had a substantial impact on the way in which the learning processes are organised.

For reasons that will be explained, the conceptual framework that has been developed within this research programme is called {\em information compression by multiple alignment, unification and search} (ICMAUS).

The next section describes the origins and motivation for the ICMAUS approach and how it relates to other work in machine learning. This is followed by a section describing the ICMAUS theory in outline. The bulk of the article describes the SP70 model---a first implementation of the ICMAUS approach to learning---and shows what it can do. Limitations of the current model---and how they may be overcome---are briefly discussed.

\section{Origins, motivation and relation to other research}

Some years ago, I developed a computer model, MK10, that is quite successful at discovering, without supervision, words and other linguistic segments in unsegmented natural language texts \citep{wolff_1980, wolff_1977, wolff_1975}. MK10 was further developed into SNPR, a model that is able, without supervision, to discover plausible context-free phrase-structure grammars from unsegmented samples of artificial languages (\citet{wolff_1982, wolff_1988}, see also \citet{langley_stromsten_2000}).

Each of these models may be seen to be largely a process of information compression (IC) by matching patterns against each other and merging or `unifying' patterns that are the same. Both models incorporate a process of searching through the abstract space of alternative matches to find those that yield relatively large amounts of compression.

To a large extent, both models operate in accordance with principles of Minimum Length Encoding (MLE) described in Section \ref{MLE_principles}, below.\footnote{MLE is an umbrella term for Minimum Message Length (MML) encoding, Minimum Description Length (MDL) encoding and related ideas such as `Kolmogorov complexity' \citep[see][]{li_vitanyi_1997}.}

In the early 1980s, I was struck by the parallels that seemed to exist between my two programs for unsupervised learning and the Prolog system, designed originally for theorem proving. A prominent feature of both learning models is IC by pattern matching, unification and search. Although IC is not a recognised feature of Prolog, a process of searching for patterns that match each other is fundamental in that system and the merging of matching patterns is an important part of `unification' as that term is understood in logic. It seemed possible that IC might have the same fundamental role in logic---and perhaps also in `computing' and mathematics---as it does in grammatical inference.
 
These observations led to the thought that it might be possible to integrate unsupervised learning and logical inference within a single system, dedicated to IC by pattern matching, unification and search. Further thinking suggested that the scope of this integrated system might be expanded to include such things as best-match information retrieval, fuzzy pattern recognition, parsing and production of language, probabilistic inference and other aspects of AI.

Some of this thinking is described in \citet{wolff_1993}.

\subsection{Development}

Development of these ideas has been underway since 1987.

Given the overall goal of integrating diverse functions within a single system, it was evident quite early that the new system would need to be organised in a way that was rather different from the organisation of the MK10 and SNPR models. And, notwithstanding the development of Inductive Logic Programming, it seemed that Prolog, in itself, was not suitable as a vehicle for the proposed developments---largely because of unwanted complexity in the system and because of the relative inflexibility of the search processes in Prolog. It seemed necessary to build the proposed new integrated system from new and `deeper' foundations.

Initial efforts focussed on the development of an improved version of `dynamic programming' for finding full matches and good partial matches between pairs of patterns \citep[see, for example,][]{sankoff_kruskall_1983}. About 1994, it became apparent that the scope of the system could be greatly enhanced by replacing the concept of `pattern matching' with the more specific concept of `multiple alignment', similar to that concept in bio-informatics but with important differences.

Notwithstanding the origin of these ideas in research on inductive learning, most of the effort to date has been focussed on such things as probabilistic reasoning \citep{wolff_2001_igpl, wolff_1999_prob}, natural language processing \citep{wolff_2000}, and the application of these ideas to the interpretation of `computing' \citep{wolff_1999_comp} and concepts in mathematics and logic \citep{wolff_maths_logic}. Now, with the integration of learning capabilities within the new framework, the circle is largely complete.

\subsection{Goals of the research}\label{goals_of_research}

All of the goals that were set for the earlier research on the learning of natural languages have been carried forward into the ICMAUS programme. The main reason for this is that the unsupervised learning of a natural language remains one of the biggest challenges for research in machine learning and it seems likely that insights gained in that area will prove useful in other areas.

By no means all the goals have been met with the current model but long-term goals that have yet to be met may, nevertheless, have a bearing on design decisions made in current work.

\subsubsection{Representation of knowledge}\label{repr_knowledge_goal}

The system for representing knowledge should be `powerful' enough to accommodate the kinds of structures found in natural languages, both syntax and semantics---and their integration.

The system should represent structures in a form that is readable and comprehensible---which means the explicit, symbolic style associated with linguistics rather than the implicit style of most artificial neural networks.

\subsubsection{The goals of learning}\label{goals_of_learning}

With respect to learning, the system should be able to achieve the following things:

\begin{itemize}

\item It should be able to learn the kinds of segmental structures found in natural languages such as words, phrases and sentences without the need for these things to be marked explicitly in the raw data.

\item It should be able to learn word classes and other distributionally-equivalent groupings of segmental structures (e.g., `nouns', `verbs' and `adjectives').

\item The learning of segmental structures and distributional classes should be integrated and it should be possible to learn both kinds of structure through two or more levels of abstraction.

\item Ultimately, the system should be able to learn such things as discontinuous dependencies in syntax, it should be able to learn semantic structures and it should be able to learn structures that integrate syntax with semantics.

\item The system should be able to create rules that generalise the structures found in the corpus of raw data, and it should be able to correct `over' generalisations that may be formed. This corresponds to children's ability to generalise beyond the language that they hear and to correct the kinds of overgeneralisations that they make when they are very young.

\item The system should be able to learn `correct' grammars despite corruption of the corpus of raw data. This corresponds to the way in which we can learn to distinguish sharply between utterances that belong in our native language and those that do not, despite the fact that the language we hear as children often contains false starts, incomplete sentences and other ungrammatical forms.

\end{itemize}

Apart from the learning of discontinuous dependencies and semantic structures, all of these goals have already been met to some degree by the SNPR model \citep{wolff_1988, wolff_1982}. In current work, the challenge has been to achieve these goals in the new framework.

\subsubsection{Unsupervised learning}

The weight of available evidence is that a child can learn his or her first language or languages without error correction by a `teacher', without the provision of `negative' samples (marked as `wrong'), and without the grading of language samples from simple to complex. In short, children can apparently learn without `supervision' although they may take advantage of error correction and the like if it is available.

Unless one subscribes to a Chomskyan `nativist' view of language learning, this feature of children's learning appears to conflict with Gold's \citeyearpar{gold_1967} proof that language learning, conceived as ``language identification in the limit'', is not possible without at least one of the sources of error correction mentioned above.

This apparent conflict can be resolved if it is accepted that Gold's concept of grammatical inference is not a good model of the way a child learns his or her native language. Rather than suppose that there is a `correct' grammar which is the `target' to be learned \citep[see also][]{valiant_1984}, we may suppose that learning is a process of optimisation. 

Given that the abstract `space' of potential grammars for any natural language is astronomically large, learning must rely on heuristic methods such as hill climbing. This means that it is not possible to know whether or not one has found the `best' grammar in terms of one's chosen criterion. However, for all practical purposes, this does not matter. All that is necessary is to find a grammar that is `good enough'.

In the light of these considerations, this project has aimed to develop a learning system conceived as an {\em unsupervised} process of optimisation, without a target grammar.

\subsubsection{Minimum Length Encoding}

A major goal in this research has been to develop the idea that MLE principles (described in Section \ref{MLE_principles}) might serve as a unifying foundation for all kinds of information processing. Care has been taken to ensure that the ICMAUS framework should operate in accordance with those principles.

\subsubsection{Integration of learning with other AI functions}

As already noted, a major goal of this research is the {\em integration} of learning with other AI functions such as pattern recognition, reasoning, planning and problem solving, and others. This requirement has led to a substantial reorganisation of learning processes compared with the earlier models (MK10 and SNPR) that were dedicated purely to learning.

\subsubsection{Realisation of the model in terms of neural mechanisms}

The ICMAUS framework is intended as an abstract model of information processing in systems of all kinds, both artificial and natural. Given that it should be relevant to understanding brains and nervous systems as well as computers, the model has been developed with an eye to its possible realisation in terms of neural mechanisms. Recent work has shown that the main elements of the model may indeed be instantiated in those terms \citep{wolff_icmaus_neural}.

\subsection{How the research relates to other research on machine learning}

The way in which this research relates to other work on machine learning is best seen in terms of the goals of the research, just described. It is, for example, more closely related to research on grammar induction using phrase-structure grammars (augmented or otherwise) than it is to research on the learning of DFAs, n-grams or Markov models---which are known to be less adequate for representing the structure of natural languages \citep{chomsky_1957}. The research is closer in spirit to other research on unsupervised learning than it is to research on learning with external error correction or negative samples. There is a closer relation with work in which learning is conceived as a process of optimisation using MLE principles than with research on ``language identification in the limit'' or related ideas. And there is a stronger affinity with systems that are designed to learn explicit symbolic structures that are readable and comprehensible than with systems---such as many artificial neural networks---that learn knowledge in an implicit form.

Studies that are, perhaps, most closely related to the present research include: \citet{adriaans_et_al_2000, allison_wallace_yee_1992, clark_2001, denis_2001, henrichsen_2002, johnson_reizler_2002, klein_manning_2001, nevill-manning_witten_1997, oliveira_sv_1996, rapp_et_al_1994, watkinson_manandhar_2001}. 

Compared with all other work on unsupervised learning of grammar-like structures, the most distinctive features of this research are:

\begin{itemize}

\item The integration of learning with other areas of AI and computation.

\item The multiple alignment concept as it has been developed in the ICMAUS framework, described below.

\end{itemize}

\section{Information Compression by Multiple Alignment, Unification and Search (ICMAUS)}\label{ICMAUS_section}

The ICMAUS framework is intended as an abstract model of any kind of system for computing or cognition, either natural or artificial. The overall organisation of the framework is as follows:

\begin{itemize}

\item The system receives raw data from the world via its `senses'. These data are designated `New'. In the course of learning, New information is transferred to a repository of stored information, initially empty, which is designated `Old'.

\item The system tries to compress each section of New as much as possible by searching for matching patterns both within the given section and between that section and information already stored in Old. If for example, the pattern `information compression' appears two or more times, it may be assigned a relatively short identifier or `code' (e.g., ``IC'') which may then be used as an abbreviation for that pattern wherever it appears.

\item In Old, the system stores repeating patterns of this kind. It also stores unmatched portions of New in case they may prove useful in compressing sections of New that come later. As we shall see, the system also creates and stores patterns that represent higher levels of abstraction. Each of the patterns stored in Old has appropriate code symbols.

\item Periodically, the patterns in Old are evaluated to differentiate those that are proving useful in the encoding of New from those that are not. The latter may be purged from the system.

\end{itemize}

In broad terms, this incremental scheme is similar to the well-known and widely-used Lempel-Ziv algorithms for IC. What is different about the ICMAUS scheme (as it has been developed for AI applications) is an emphasis on partial matching and on relatively thorough searching of the space of alternative possible matches. Also distinctive is the concept of `multiple alignment' as it has been developed in this research to support the encoding of New information in terms of Old information in a hierarchy of `levels' (as will be seen in examples below).

Information compression may be interpreted as a process of maximising {\em Simplicity} in information (by removing redundancy) whilst retaining as much as possible of its non-redundant, descriptive {\em Power}. Hence the sobriquet `SP' that has been adopted for the ICMAUS proposals and the computer models in which the framework is realised.

\subsection{Representation of Knowledge}\label{representation_of_knowledge}

Given the intended wide scope of the framework, a goal of the research has been to devise a `universal' scheme for the representation of knowledge, capable of representing diverse kinds of knowledge in a succinct manner and capable of integrating diverse kinds of knowledge in a seamless manner. Naturally, the design of the scheme would depend, in part, on the ways in which stored knowledge is going to be used.

What has emerged is almost the simplest conceivable format for knowledge: arrays or {\em patterns} of atomic symbols in one or more dimensions. So far, the focus has been on one-dimensional arrays---`strings' or `sequences' of symbols. However, since the system is intended eventually to embrace arrays in two dimensions and possibly more, the relatively general term `pattern' is normally used.

In this context, a `symbol' is merely a `mark' that can be discriminated in a yes/no manner from other symbols. In all the examples in this paper, each symbol is represented by a string of one or more non-space characters bounded by spaces.

In general, symbols do not have `hidden' meanings (e.g., ``multiply'' for the symbol `$\times$')---any meanings that may attach to symbols within a given body of knowledge are to be expressed in the form of other symbols and patterns within that knowledge.

An apparent exception to the slogan ``no hidden meanings'' is that, within each pattern, there is normally a distinction between symbols that represent the substance or contents of the pattern and other symbols that serve to identify or `encode' the pattern (as will be described). This kind of distinction may be justified on the grounds that it is part of the mechanism by which the system organises and uses its knowledge. What the slogan really applies to are meanings that are part of the knowledge itself.

The `granularity' of symbols in the ICMAUS framework is undefined. Symbols may be used to represent very fine-grained details of a body of information or they may be used to represent relatively large chunks of information.

Despite the extreme simplicity of the format for representing knowledge, the way it is processed within the ICMAUS framework means that it can model a variety of established representational schemes including context-free phrase-structure grammars (CF-PSGs), context-sensitive grammars, production rules, networks, trees and other schemes. Some examples will be seen below and many more may be found in \citet{wolff_icmaus_overview, wolff_maths_logic, wolff_2000, wolff_1999_comp, wolff_1999_prob}.

With respect to the representation of structures in NL (Section \ref{repr_knowledge_goal}), the ICMAUS system can accommodate the major features of NL syntax, including discontinuous dependencies \citep{wolff_2000}, it can represent various kinds of non-syntactic `semantic' structures \citep[see][]{wolff_icmaus_overview} and preliminary work suggests that syntax and semantics can indeed be integrated.

\subsection{Minimum Length Encoding}\label{MLE_principles}

The ICMAUS framework is founded explicitly on MLE principles pioneered by \citet{solomonoff_1997, solomonoff_1964} and also by \citet{wallace_boulton_1968} and \citet{rissanen_1978} \citep[see also][]{li_vitanyi_1997}. 

The key idea in MLE is that, in grammatical inference and related kinds of learning, one should seek to minimise $(G + E)$, where $G$ is the size of the grammar (in bits or equivalent measure) and $E$ is the size of the sample (in bits) after it has been encoded in terms of the grammar. This principle guards against the induction of grammars that are trivially small (where a small $G$ is offset by a disproportionately large $E$) and it also guards against the induction of grammars where a small value for $E$ is achieved at the cost of a disproportionately large value for $G$.

\subsubsection{MLE and Human Intuition}\label{MLE_and_intuition}

Part of the motivation for focussing on MLE principles is that they seem to have a bearing on human perception and learning \citep[see][]{chater_1999, chater_1996, wolff_1988} and the human brain is currently the best learning system in existence. Human intuitions about what constitutes `good' or `natural' ways of structuring information do seem to seem to be conditioned to a large extent by MLE principles \citep{wolff_1988, wolff_1977}.

\subsubsection{MLE and Bayesian principles}\label{MLE_Bayesian}

It is widely recognised that MLE principles may be translated into Bayesian concepts of probability \citep[see, for example,][]{eisner_2002}. Although the two views are equivalent at one level of analysis, they differ in terms of their heuristic value: primitive concepts of pattern matching and the unification of identical patterns have a much more direct and transparent relation to IC than they do to probabilistic concepts. In this research, a focus on IC by the matching and unification of patterns has proved to be a more fruitful source of new insights than analyses in terms of probabilities.

\subsection{Computer Models}

During the development of the ICMAUS framework, computer models have been developed as a means of testing the ideas and also as a means of demonstrating what can be done with the framework.

Two main models have been developed to date:

\begin{itemize}

\item SP61 which realises the encoding of New information in terms of pre-defined Old information but without any attempt to modify Old by adding patterns or purging them. SP61 also contains procedures for calculating probabilities of inferences that may be drawn by the system, as described in Section \ref{probabilistic_reasoning}, below.

\item SP70 which realises all four elements of the framework, including the addition of patterns to Old and their evaluation in terms of MLE principles.

\end{itemize}

Although SP61 is largely a subset of SP70, it is convenient to run it as a separate model for some applications. The organisation of SP70 is described in Section \ref{SP70_section}, below.

\subsection{Applications}

As an indication of the scope of the ICMAUS framework, this subsection briefly sketches some of the things that it can do. 

The first example (parsing of natural language) is described relatively fully because it serves to illustrate the main elements of the multiple alignment concept as it has been developed in this research. Owing to limitations of space, the remaining examples of ICMAUS applications are necessarily brief.

A much fuller overview of the framework and its capabilities is provided in \citet{wolff_icmaus_overview} and more detail about specific aspects of the framework may be found in other sources cited below.

\subsubsection{Natural Language Processing}\label{NL_processing}

Figure \ref{alignment_1} shows the best alignment found by SP61 with the sentence `o n e o f t h e m d o e s' in New and patterns representing grammatical rules in Old. In this context, `best' means the alignment that allows New to be encoded economically in terms of patterns in Old, as explained below. 

By convention, the New pattern is always shown at the top of each alignment with patterns from Old in the rows underneath, one in each row. The order of the rows below the top row is entirely arbitrary and has no special significance.

\begin{figure}[!hbt]
\fontsize{06.00pt}{07.20pt}
\begin{center}
\begin{BVerbatim}
0                             o n e               o f            t h e m                d o e s   0
                              | | |               | |            | | | |                | | | |  
1                             | | |               | |   < N Np 0 t h e m >              | | | |   1
                              | | |               | |   | |              |              | | | |  
2                             | | |   < Q 0 < P   | | > < N              > >            | | | |   2
                              | | |   | |   | |   | | |                    |            | | | |  
3                             | | |   | |   < P 2 o f >                    |            | | | |   3
                              | | |   | |                                  |            | | | |  
4                             | | |   | |                                  |   < V Vs 1 d o e s > 4
                              | | |   | |                                  |   | | |            |
5 S Num     ; < NP            | | |   | |                                  | > < V |            > 5
     |      | | |             | | |   | |                                  | |     |             
6    |      | | |    < N Ns 3 o n e > | |                                  | |     |              6
     |      | | |    | | |          | | |                                  | |     |             
7    |      | < NP 0 < N |          > < Q                                  > >     |              7
     |      |            |              |                                          |             
8   Num SNG ;            Ns             Q                                          Vs             8
\end{BVerbatim}
\end{center}
\caption{The best alignment found by SP61 with `o n e o f t h e m d o e s' in New and patterns representing grammatical rules in Old.}
\label{alignment_1}
\end{figure}

Apart from the pattern in row 8, the patterns from Old in this example are like re-write rules in a context-free phrase-structure grammar (CF-PSG) with the re-write arrow omitted. If we ignore row 8, the alignment shown in Figure \ref{alignment_1} is very much like a conventional parsing, marking the main components of the sentence: words and phrases and the sentence pattern itself (shown in row 5).

Row 8 shows how the `discontinuous' dependency that exists between the singular noun in the subject of the sentence (`Ns') and the singular verb (`Vs') can be marked within the alignment in a relatively direct manner. Despite the simplicity of the format for representing knowledge, the formation of multiple alignments enables the system to express `context sensitive' aspects of language and other kinds of knowledge.

Within each pattern, some symbols have the status of {\em identification} (ID) symbols and others serve as {\em contents} (C) symbols. In general, ID symbols in any given pattern comprise bracket symbols (`$<$' and '$>$') that define the left and right boundaries of the pattern (where that is necessary) together with one or more symbols, usually near the start of the pattern, that serve to identify the pattern uniquely within Old or otherwise define how it relates to other patterns. All other symbols are C symbols. Examples of ID symbols in Figure \ref{alignment_1} include `$<$ N Np 0 $>$' in row 1, `$<$ NP 0 $>$' in row 7 and `Num SNG ; Q' in row 8. The corresponding C symbols are `t h e m' in row 1, `$<$ N $>$ $<$ Q $>$' in row 7 and `Ns Vs' in row 8.\footnote{Notice that the distinction between ID symbols and C symbols does not depend on their appearance. The status of each symbol within a pattern is marked when the pattern is created.}

An encoding for New can be derived from each alignment by looking for columns in the alignment containing a single ID symbol, not matched to any other symbol, and copying these symbols into a {\em code pattern} in the same sequence as they appear in the alignment. In this example, the code pattern derived in this way is `S SNG 0 3 0 0 1'.

A neat feature of the ICMAUS framework is that, {\em without any modification}, it may be used to produce sentences as well as parse them. If the program is run again with the sentence in New replaced by the code pattern `S SNG 0 3 0 0 1', the best alignment found by the system is, apart from the top row, the same as the one shown in Figure \ref{alignment_1}. All the words in the original sentence appear within the alignment in their correct order.

Much more detail, with many more examples, may be found in \citet{wolff_2000}.

\subsubsection{Fuzzy Pattern Recognition and Best-Match Information Retrieval}

The dynamic programming built into the SP models means that they are just as at home with `fuzzy', partial matches between patterns as with exact matches. This means they lend themselves well to modelling human-like capabilities for recognising patterns and objects despite errors of omission, commission or substitution, including the way in which we can recognise something even if it is partially obscured by other things. In a similar way, the models are able to model best-match retrieval of information from Old.

Examples may be seen in \citet{wolff_icmaus_overview, wolff_2001_igpl, wolff_1999_prob}.

\subsubsection{Class Hierarchies with Inheritance of Attributes}

The ICMAUS framework also lends itself quite well to modelling class hierarchies with inheritance of attributes and recognition of objects or patterns at multiple levels of abstraction. Any class and its associated attributes may be represented by a pattern and `isa' links between classes may be established in much the same way as grammatical patterns are linked by means of matching symbols in the parsing example shown in Figure \ref{alignment_1}.

Examples may be seen in \citet{wolff_icmaus_overview, wolff_1999_prob}.

\subsubsection{Probabilistic Reasoning}\label{probabilistic_reasoning}

Any column within an alignment that does {\em not} contain a symbol from New may be regarded as an inference that may be drawn from the alignment. For example, the column in Figure \ref{alignment_1} that contains (two instances of) the symbol `V' may be seen as representing the inference that `d o e s' is a verb.

This and related aspects of the ICMAUS framework provide a powerful means of modelling several kinds of probabilistic inference including probabilistic chains of reasoning, abductive reasoning, default reasoning, nonmonotonic reasoning, the phenomenon of ``explaining away'' and the solution of geometric analogy problems. These aspects of the ICMAUS framework are explained quite fully, with examples, in \citet{wolff_1999_prob} \citep[see also][]{wolff_2001_igpl}.

\subsubsection{Modelling `Computing' and Concepts in Mathematics and \newline Logic}

In \citet{wolff_1999_comp}, I have argued that the ICMAUS framework provides an interpretation for the Turing model of `computing' and equivalent models such as the Post Canonical System. Although the ICMAUS model is a little more complex than earlier models it appears to illuminate a range of issues (mainly in AI) that are outside the scope of earlier models.

In \citet{wolff_maths_logic}, I have argued that the ICMAUS framework provides an interpretation for a range of concepts in mathematics and logic, including both the static structures found in those disciplines and the dynamics of calculation and inference.

\section{SP70}\label{SP70_section}

All the main components of the ICMAUS framework outlined in Section \ref{ICMAUS_section} are now realised within the SP70 software model (version 9.2).

In the description of the model that follows, the examples have a `linguistic' flavour with an emphasis on the simpler aspects of English syntax. However, the term `grammar' in this context should be construed very broadly since, as we noted in Section \ref{representation_of_knowledge}, it assumed that similar principles may be applied to many different kinds of knowledge.

The model gives results in the area of unsupervised learning that are good enough to show that the framework is sound. As we shall see, the model is able to abstract plausible grammars from sets of simple sentences without prior knowledge of word segments or the classes to which they belong (Section \ref{examples_section}) and the computational complexity of the model appears to be acceptable (Section \ref{computational_complexity}).

However, in its current form, the model has at least two significant shortcomings and some other deficiencies. A programme of further development, experimentation and refinement is still needed to realise the full potential of the model for unsupervised learning. 

This model is governed by mathematical principles---explained at pertinent points below---but these are remarkably simple and in accordance with established theory. The main focus in what follows is on the organisation of the model and the computational techniques employed within it.

\subsection{Objectives}

In keeping with the goals of learning described above (Section \ref{goals_of_learning}), the main problems addressed in the development of this model have been:

\begin{itemize}

\item How to identify significant segments in the `corpus' of data which is the basis of learning when the boundary between one segment and the next is not marked explicitly.

\item How to identify classes of syntactically-equivalent segments.

\item How to combine the learning of segmental structure with the learning of disjunctive classes.

\item How to learn segments and disjunctive classes through two or more levels of abstraction.

\item \sloppy How to generalize grammatical rules beyond the data and how to correct over-generalizations without feedback from a `teacher' or the provision of `negative' samples or the grading of the data from `easy' to `hard'.

\end{itemize}

Solutions to these problems were found in the SNPR model \citep{wolff_1988, wolff_1982} but, as noted earlier, the organisation of this model is quite unsuited to the wider goals of the present research---integration of diverse functions within one framework. Finding new solutions to these problems within the ICMAUS framework has been a significant challenge.

The SP70 model (v.~9.2) provides solutions to the first three problems and partial solutions to the fourth and fifth problems. Further development will be required to achieve robust learning of structures with more than two levels of abstraction and more work is required on the generalization of grammatical rules and the correction of overgeneralizations (see Section \ref{discussion}, below).

\subsection{Overall Structure of the Model}

Figure \ref{SP70_figure} shows the high-level organisation of the SP70 model. The program starts with a set of New patterns and with a repository of Old patterns that is initially empty.

In broad terms, the model comprises two main phases:

\begin{itemize}

\item Create a set of patterns that may be used to encode the patterns from New in an economical manner (operations 1 to 3 in Figure \ref{SP70_figure}).

\item From the patterns created in the first phase, compile one or more alternative `grammars' for the patterns in New in accordance with MLE principles (operation 4 in Figure \ref{SP70_figure}).

\end{itemize}

After reading a set of patterns into New, the system compiles an `alphabet' of the different types of symbols appearing in those patterns, counts their frequencies of occurrence and calculates encoding costs as described below. These values will be needed for the evaluation of alignments. The term {\em symbol type} in this connection means a representative template or example of a set of identical symbols.

\begin{figure}[!hbt]
\fontsize{08.00pt}{09.60pt}
\begin{center}
\begin{BVerbatim}
SP70()
{
     1 Read a set of patterns into New. Old is initially empty.
     2 Compile an alphabet of symbol types in New and, for each type,
          find its frequency of occurrence and the number of bits
          required to encode it.
     3 While (there are unprocessed patterns
          in New)
     {
          3.1 Identify the first or next pattern from New as the
               `current pattern from New' (CPFN).
          3.2 Apply the function CREATE_MULTIPLE_ALIGNMENTS() to
               create multiple alignments, each one between the
               CPFN and one or more patterns from Old.
          3.3 During 3.2, the CPFN is copied into Old, one symbol
               at a time, in such a way that the CPFN can be
               aligned with its copy but that any one symbol in
               the CPFN cannot be aligned with the corresponding
               symbol in the copy.
          3.4 Sort the alignments formed by this function in order
               of their compression scores.
          3.5 From amongst the best alignments, select a subset
               that conforms to other constraints described in the
               text.
          3.6 Process the selected alignments with the function
               DERIVE_PATTERNS(). This function derives encoded
               patterns from alignments and adds them to Old.
     }

     4 Apply the function SIFTING_AND_SORTING() to create one or
          more alternative grammars for the patterns in New, each
          one scored in terms of MLE principles. Each grammar is
          a subset of the patterns in Old.
}
\end{BVerbatim}
\end{center}
\caption{The organisation of SP70. The workings of the functions {\em create\_multiple\_alignments()}, {\em derive\_patterns()} and {\em sifting\_and\_sorting()} are explained in Sections \ref{create_multiple_alignments_section}, \ref{derive_patterns_section} and \ref{sifting_and_sorting_section}, respectively.}
\label{SP70_figure}
\end{figure}

Next (operation 3), each pattern from New is processed by searching for alignments that allow the given pattern to be encoded economically in terms of patterns in Old (as outlined in Section \ref{NL_processing}). From a selection of the best alignments found, the program `learns' new patterns, as explained below, and adds them to Old. A copy of each pattern from New is also added to Old, marked with new ID symbols as will be explained.

When all the patterns from New have been processed in this way, there is a process of sifting and sorting to create one or more alternative grammars for the patterns from New (operation 4). Each grammar comprises a subset of the patterns in Old and each one is scored in terms of MLE principles.

\subsection{Creating Multiple Alignments}\label{create_multiple_alignments_section}

The function {\em create\_multiple\_alignments()} referred to in Figure \ref{SP70_figure} creates zero or more multiple alignments, each one comprising the current pattern from New (CPFN\footnote{Where the word ``current'' is not required, a pattern from New will be referred as a PFN.}) and one or more patterns from Old. Each alignment has a {\em compression score} which is a measure of the amount of compression of the CPFN that can be achieved by encoding it in terms of patterns from Old. 

Apart from some minor modifications and improvements, this function is essentially the same as the main component of the SP61 model, described quite fully in \citet{wolff_2000}. Readers are referred to this source for a more detailed description of how multiple alignments are formed in the ICMAUS framework.

At the heart of the function is a process for finding full alignments and good partial alignments between pairs of patterns. This process, described quite fully in \citet{wolff_1994_scaleable}, is a version of `dynamic programming' \citep[see][]{sankoff_kruskall_1983} with advantages over standard methods:

\begin{itemize}

\item List processing techniques allow memory to be used efficiently so that long patterns may be matched.

\item In cases where a pair of patterns can be matched in more than one way, the process can deliver two or more alternative alignments, each one scored in terms of compression.

\item {\sloppy The `depth' or thoroughness of searching---and thus the speed of \newline processing---can be varied.}

\end{itemize}

This matching process is applied iteratively to build alignments containing two or more rows, one pattern per row. The process stops when no more alignments can be found or when compression scores have reached a peak and have fallen for two or three cycles.

On the first cycle, the matching process is applied to find alignments, each one between the CPFN and {\em one} of the patterns stored in Old. When this has been completed, they are sorted in order of their compression scores and several of the best are selected for further processing. In general, the alignments that are selected are ones that can be treated as if they were a single sequence (pattern) of symbols, as described in Section \ref{constraints_2}, below.

In the second and subsequent cycles of the building process, two or three of the best alignments selected in the previous cycle are chosen as `driving' patterns and matched against `target' patterns comprising the current set of patterns in Old together with all of the alignments selected in previous cycles (including the driving patterns themselves). Each of the resulting alignments is between one driving pattern and one target pattern. As on the first cycle, several of the best alignments formed are selected for further processing provided they can be treated as if they were single sequences of symbols.

\subsubsection{Calculation of Compression Scores}\label{calculation_of_compression_scores}

The compression score for an alignment is calculated as:

\[C = N_r - N_e\]

\noindent where $N_r$ is the size, in bits, of the CPFN in its uncompressed `raw' form and $N_e$ is the size, in bits, of the code pattern derived from the alignment as indicated in Section \ref{NL_processing}. $N_r$ is calculated as:

\[N_r = \sum_{i = 1}^{i = n} s_i\]

\noindent where $s_i$ is number of bits required to encode the $i$th symbol in the sequence of $n$ symbols in the CPFN. $N_e$ is calculated in the same way from the symbols in the code pattern.

With a qualification to be described, the number of bits required to encode any given symbol type that appears in New (its `encoding cost') is calculated using the Shannon-Fano-Elias (SFE) method \citep[see][]{cover_thomas_1991}. This method is similar to the well-known Huffman coding method and gives similar results---but it has advantages over the Huffman method when codes are used for the calculation of probabilities \citep[see][]{wolff_1999_prob}. As noted above, the values for the frequencies of symbol types that are required for this method are computed (in operation 2 in Figure \ref{SP70_figure}) from the set of patterns in New.

Notice that the encoding cost of any symbol is totally independent of the size of the symbol as it appears to the reader. In general, symbols are represented by character strings chosen for reasons of readability or mnemonic value. These strings are quite independent of the number of bits required to discriminate one symbol from another in an efficient manner.

The qualification mentioned above is that the encoding costs of symbol types appearing in New (calculated by the SFE method) are multiplied by a {\em cost factor}, normally about 5 or 10. This means that symbols representing the original raw `data' for the program are treated by the system as if they were relatively large chunks of information---by contrast with other symbols that are used to encode the data (added by the system in the course of learning) which are not given any additional weighting. The reason for applying this cost factor to data symbols is that, without this weighting, the best grammar found for the kind of small example that is convenient for experimentation and demonstration is often simply a repetition of the original patterns, without any recognition of structures within those patterns. If the data symbols are treated as if they were larger chunks of information, then the benefits of recognising substructures outweighs the costs of encoding those structures. With larger examples, where frequency values for substructures will normally be higher, this problem should disappear.

As learning proceeds, patterns are added to Old (operation 3.6 in Figure \ref{SP70_figure}). Within those patterns, some of the symbols are derived from New and their encoding costs are already known. However, other symbols---ID symbols and copies of them---are created by the system in operation 3.6 (Figure \ref{SP70_figure}) and the variety of types of these symbols and their frequencies of occurrence are constantly changing. For this reason, it is difficult at this stage to calculate encoding costs using the SFE method. Accordingly, the encoding costs of symbols created by the system are initially set at a fixed arbitrary value. As we shall see, more precise values are calculated in the {\em sifting\_and\_sorting()} phase of processing. The approximation at this stage does not seem to be a serious impediment to learning, perhaps because the selection of alignments depends on relative values for compression scores, not absolute values.

The search for alignments that maximise $C$ conforms with MLE principles as outlined in Section \ref{MLE_principles}. The repository of Old patterns may be taken to be the current `grammar' for encoding the CPFN and, for any given CPFN, the size of this grammar, $G$, is constant.\footnote{As indicated in operation 3.3 of Figure \ref{SP70_figure} and described in Section \ref{copying_CPFN}, below, a copy of the CPFN is added to Old during the process of building alignments. However, for the purpose of encoding the CPFN, the entire copy (with ID symbols that are added to the copy) counts as part of Old and thus, for any given CPFN, $G$ is indeed constant.} Since $G$ is constant, the goal of minimising $T$ is equivalent to a goal of minimising $E$. For any given CPFN, $E$ is the same as $N_e$. Thus, since $N_r$ is constant for any given CPFN, seeking to minimise $E$ is equivalent to the attempt to maximise $C$.

\subsubsection{Constraints on Matching when a Given Pattern Appears Two or More Times in an Alignment}\label{constraints_1}

An important point to notice about multiple alignments in the ICMAUS framework is that any pattern may appear two or more times in one alignment. For example, in a sentence like {\em The winds from the west are strong}, there are two instances of a simple form of noun phrase ({\em the winds} and {\em the west}) so that, in a multiple alignment parsing of the sentence, there would be two appearances of the pattern representing the structure of that form of noun phrase.

Notice that multiple {\em appearances} of one pattern within an alignment are {\em not} the same as multiple {\em copies} of the pattern within an alignment. In the latter case, there are two or more distinct patterns and it is quite acceptable for them to be fully aligned, one with another. In the former case, there is only a single pattern so it is not permissible for a symbol in one appearance to be matched against the corresponding symbol in another appearance---because this would mean matching the given symbol with itself. However, it is permissible to form a match between one symbol within the given pattern and another symbol (in another position) in the same pattern.

In any situation where a given pattern is matched---directly or indirectly---with itself, the matching process in the {\em create\_multiple\_alignments()} function is constrained to prevent any one symbol being matched with itself.

\subsubsection{Constraints on ``Mismatches'' between Patterns and the Treatment of Alignments as Simple Sequences}\label{constraints_2}

A {\em mismatch} in an alignment occurs where one or more unmatched symbols in one pattern appear opposite one or more unmatched symbols in another pattern within the alignment. Symbols are `opposite' each other if they lie between two columns of matched symbols or between one column of matched symbols and the beginning or end of the alignment. 

In the ICMAUS framework, mismatches are illegal if they occur {\em between patterns from Old}. If any alignment contains that kind of mismatch, it is discarded. Notice that a mismatch between the CPFN and any pattern from Old is legal and, as we shall see, it is those kinds of mismatches (and, more generally, unmatched C symbols within alignments) that drive the learning process.

In each of the alignments shown in Figure \ref{mismatch_figure}, row 0 contains a pattern from New and rows 1 and 2 contain patterns from Old. Where `x' and `y' lie opposite each other (in (a) and (b)), there is an illegal mismatch. Where `b' or `c' lies opposite `x' or `y' (or both) (in all four alignments), the mismatch is legal.

\begin{figure}[!hbt]
\begin{center}
\begin{BVerbatim}
0 a b c 0             0 a b c 0
  |   |                 | |
1 a x c 1             1 a b x 1
  |   |                 | |
2 a y c 2             2 a b y 2

(a)                   (b)

0 a b c 0             0 a b c 0
  |   |                 | |
1 a x c 1             1 a b   1
  |   |                 | |
2 a   c 2             2 a b y 2

(c)                   (d)
\end{BVerbatim}
\end{center}
\caption{Alignments illustrating mismatches as discussed in the text.}
\label{mismatch_figure}
\end{figure}

As was noted above, the process of building multiple alignments requires that each alignment created in the intermediate stages can be treated as a single sequence of symbols. The critical issue here is whether or not a given alignment contains any illegal mismatches. If it does contain illegal mismatches, it cannot be treated as a single sequence. Otherwise, it can. Notice that mismatches between the CPFN and other patterns are of no consequence. For example, alignment (c) in Figure \ref{mismatch_figure} may be treated as the sequence `a x c' while alignment (d) in the same figure may be treated as `a b y'. All unmatched symbols within the CPFN are simply ignored.

\subsection{Copying the CPFN into Old}\label{copying_CPFN}

In its bare essentials, `learning' in SP70 is achieved by the addition of patterns to Old. This occurs in two ways: by copying each pattern from New into Old (operation 3.3 in Figure \ref{SP70_figure}) and by deriving patterns from alignments in the function {\em derive\_patterns()} (operation 3.6 in the same figure). The first of these is described here and the second is described in the next subsection.

{\sloppy During the matching process in the first cycle of the \newline {\em create\_multiple\_alignments()} function, the CPFN is copied, one symbol at a time, into Old in such a way that any symbol in the CPFN can be matched with any earlier symbol in the copy but it cannot be matched with the corresponding symbol in the copy or any subsequent symbol. When the transfer is complete, ID symbols are added to the copy to provide a `code' for the pattern, as described below.}

The aim here is to detect any redundancy that may exist {\em within} each pattern from New (e.g., the repetition that can be seen in the pattern `a b c d x y z a b c d') but to avoid detecting the redundancy resulting from the fact that the CPFN has been copied into Old. This constraint is imposed for very much the same reason as the constraint (described in Section \ref{constraints_1}, above) which prevents any one symbol within an alignment being matched with itself.

The reason for copying each pattern from New into Old rather than simply moving it is that each such pattern (with its ID symbols) is a candidate for inclusion in one or more of the best grammars selected by the {\em sifting\_and\_sorting()} function and it cannot be evaluated properly unless it is a copy of the corresponding pattern from New, not the pattern itself (see Section \ref{sifting_and_sorting_section}, below). 

The ID symbols that are added to the copy of each CPFN comprise left and right brackets (`$<$' and `$>$') at each end of the pattern together with symbols immediately after the left bracket that serve to identify the pattern uniquely amongst the patterns in Old. For the sake of consistency with the {\em derive\_patterns()} function (see Section \ref{derive_patterns_section}, next), two ID symbols follow the left bracket. Thus, for example, a pattern from New like `t h a t b o y r u n s' might become `$<$ \%1 9 t h a t b o y r u n s $>$' when ID symbols have been added.

\subsection{Deriving Patterns from Alignments}\label{derive_patterns_section}

In operation 3.6 in Figure \ref{SP70_figure}, the {\em derive\_patterns()} function is applied to a selection of the best alignments formed and, in each case, it looks for sequences of unmatched symbols within the alignment and also sequences of matched symbols.

Consider the alignment shown in Figure \ref{alignment_2}. From an alignment like that, the function finds the unmatched sequences `g i r l' and `b o y' and, within row 1, it also finds the matched sequences `t h a t' and `r u n s'. With respect to row 1, the focus of interest is the matched and unmatched sequences of C symbols---ID symbols are ignored.

\begin{figure}[!hbt]
\begin{center}
\begin{BVerbatim}
0        t h a t g i r l r u n s   0
         | | | |         | | | |  
1 < %1 9 t h a t b o y   r u n s > 1
\end{BVerbatim}
\end{center}
\caption{A simple alignment from which other patterns may be derived.}
\label{alignment_2}
\end{figure}

A copy of each of the four sequences is made, ID symbols are added to each copy (as described in Section \ref{assigning_id_symbols}, below) and the copy is added to Old. In addition, another `abstract' pattern is made that records the sequence of matched and unmatched patterns within the alignment. The result in this case is five patterns like those shown in Figure \ref{patterns_figure_1}.

\begin{figure}[!hbt]
\begin{center}
\begin{BVerbatim}
< %7 12 t h a t >
< %9 14 b o y >
< %9 15 g i r l >
< %8 13 r u n s >
< %10 16 < %7 > < %9 > < %8 > >
\end{BVerbatim}
\end{center}
\caption{Patterns derived from the alignment shown in Figure \ref{alignment_2}.}
\label{patterns_figure_1}
\end{figure}

It should be clear that the set of patterns in Figure \ref{patterns_figure_1} is, in effect, a simple grammar for the two sentences in Figure \ref{alignment_2}, with patterns representing grammatical rules in much the same style as those shown in Figure \ref{alignment_1}. The abstract pattern `$<$ \%10 220 $<$ \%7 $>$ $<$ \%9 $>$ $<$ \%8 $>$ $>$' describes the overall structure of this kind of sentence with slots that may receive individual words at appropriate points in the pattern.

Notice how the symbol `\%9' serves to mark `b o y' and `g i r l' as alternatives in the middle of the sentence. This is a grammatical class in the tradition of distributional or structural linguistics \citep[see, for example,][]{fries_1952, harris_1951}.

With alignments like this:

\begin{center}
\begin{BVerbatim}
0        t h e g r e e n a p p l e   0
         | | |           | | | | |
1 < %1 2 t h e           a p p l e > 1 
\end{BVerbatim}
\end{center}

\noindent or this:

\begin{center}
\begin{BVerbatim}
0        t h e           a p p l e   0
         | | |           | | | | |  
1 < %1 2 t h e g r e e n a p p l e > 1
\end{BVerbatim}
\end{center}

\noindent the system derives patterns very much as before except that the unmatched sequence (`g r e e n') is assigned to a class by itself, without any alternative pattern that may appear in the same context. Arguably, there should be some kind of `null' alternative to `g r e e n' in cases like this in order to capture the idea that ``the apple'' and ``the green apple'' are acceptable variants of the same phrase. This is a possible refinement of the model in the future.

Readers may wonder why the grammar shown in Figure \ref{patterns_figure_1} was not simplified to something like this:

\begin{center}
\begin{BVerbatim}
< %9 14 b o y >
< %9 15 g i r l >
< %10 16 t h a t < %9 > r u n s >
\end{BVerbatim}
\end{center}

The main reason for adopting the style shown in Figure \ref{patterns_figure_1} is that the overall organisation of the model is simpler if each newly-derived pattern is automatically referenced from the contexts or contexts in which it may appear. Another reason is that it is anticipated that, with realistically large corpora, most of the patterns that will ultimately turn out to be significant in terms of MLE principles will appear in two or more contexts and, in that case, MLE principles are likely to dictate that each pattern should be referenced from each of its contexts rather than written out redundantly in each of the two or more places where it appears.

\subsubsection{Assignment of Identification Symbols}\label{assigning_id_symbols}

Apart from the terminating brackets, each pattern in Figure \ref{patterns_figure_1} has two ID symbols:

\begin{itemize}

\item A `class' symbol (e.g., `\%7' or `\%9') that normally starts with the `\%' character. The class symbol is, in effect, a reference to the context or contexts in which the given pattern may appear. Thus, for example, the symbol `\%7' in the first pattern in Figure \ref{patterns_figure_1} shows that that pattern may appear where the matching symbol occurs in the pattern `$<$ \%10 220 $<$ \%7 $>$ $<$ \%9 $>$ $<$ \%8 $>$ $>$'. Any one pattern may belong to more than one class and should contain a symbol for each of the classes it belongs to (see Section \ref{avoiding_duplication}).

\item A `discrimination' symbol (e.g., `12', `14') that serves to distinguish the pattern from any others that may belong in the same class. At this stage, the discrimination symbol is simply a unique identifier for the given pattern amongst the other patterns and alignments created by the program.

\end{itemize}

While alignments are being built and coded patterns are being added to Old, new class symbols and new discrimination symbols are created quite liberally. However, many of these symbols are weeded out during the {\em sifting\_and\_sorting()} phase of processing and those that remain are renamed in a tidy manner.

\subsubsection{Avoiding Duplication}\label{avoiding_duplication}

In the course of deriving patterns from alignments and adding them to Old, it can easily happen that a newly-derived pattern has the same C symbols as one that is already in Old. For this reason, each newly-derived pattern is checked against patterns already stored in Old and it is discarded if an existing pattern is found with the same C symbols.

Although the discarded pattern has the same C symbols as a pre-existing pattern, it comes from a different context. So a new symbol type is created to represent that context, a copy of the symbol type is added to the pre-existing pattern, and another copy of the symbol type is added to the abstract pattern in the appropriate position. In this way, any one sequence of C symbols may appear in a pattern containing several different class symbols, each one representing one of the contexts where the C symbols may appear.

As the program stands, there is a one-to-one relation between contexts and classes. But it can easily happen that the set of patterns that may appear in one context is the same as the set of patterns that may appear in another. At some stage, it is intended that the program will be augmented to check for this kind of redundancy and to merge classes that turn out to be equivalent. 

\subsubsection{Deriving Patterns from Alignments Containing Three or More Rows}

For any given alignment, the {\em derive\_patterns()} function works by looking for one or more unmatched sequences of symbols in the CPFN, or one or more sequences of unmatched C symbols in a pattern from Old, or both these things. What happens if an alignment contains two or more patterns from Old?

Consider the alignment shown in Figure \ref{alignment_3}. In a case like this, it is necessary to identify {\em one} of the patterns from Old for the purpose of deriving patterns from the alignment. The pattern that is chosen is the one that is deemed to be the most `abstract' pattern amongst those in the rows below the top row.

\begin{figure}[!hbt]
\fontsize{08.00pt}{09.60pt}
\begin{center}
\begin{BVerbatim}
0               t h e   r e d         a p p l e          f a l l s     0
                | | |                 | | | | |          | | | | |    
1               | | |                 | | | | |   < %4 5 f a l l s >   1
                | | |                 | | | | |   | |              |  
2 < %5 7 < %1   | | | > < %2 > < %3   | | | | | > < %4             > > 2
         | |    | | | |        | |    | | | | | |                     
3        | |    | | | |        < %3 3 a p p l e >                      3
         | |    | | | |                                               
4        < %1 0 t h e >                                                4
\end{BVerbatim}
\end{center}
\caption{An alignment (between a pattern from New and four patterns from Old) from which other patterns may be derived.}
\label{alignment_3}
\end{figure}

The most abstract row in any alignment is the row below the top row that starts furthest to the left within the alignment, e.g., row 5 in Figure \ref{alignment_1} and row 2 in Figure \ref{alignment_3}. Typically, this is also the row that finishes furthest to the right. In general, this is the row within any alignment that, directly or indirectly, encodes the largest number of symbols from the CPFN. Of course, if an alignment contains only two rows, then row 1 is the most abstract row.

To be a suitable candidate for processing by the {\em derive\_patterns()} function, the only row below the top row that may contain unmatched C symbols is the most abstract row. Also, there must be at least one unmatched C symbol somewhere within the CPFN and the most abstract row. Any alignment that does not meet these conditions is discarded for the purpose of deriving new patterns.

From the alignment shown in Figure \ref{alignment_3}, the function creates patterns like those shown in Figure \ref{patterns_figure_2} and adds them to Old. 

\begin{figure}[!hbt]
\begin{center}
\begin{BVerbatim}
< %2 9 r e d >
< %7 10 < %3 > < %4 > >
< %8 11 < %1 > < %2 > < %7 > >
\end{BVerbatim}
\end{center}
\caption{A set of patterns derived from the alignment shown in Figure \ref{alignment_3}.}
\label{patterns_figure_2}
\end{figure}

Here, the unmatched sequence `r e d' from the CPFN has been converted into the pattern `$<$ \%2 9 r e d $>$'. The system recognises that `r e d' lies opposite the sequence `$<$ \%2 $>$' within row 2 of Figure \ref{alignment_3} and that this sequence is a reference to the class `\%2'. Accordingly, the pattern `r e d' has been assigned to that class. If the unmatched sequence opposite `r e d' could not be recognised as a reference to a class, or if there was no unmatched sequence opposite `r e d', then the system would create a new class and new patterns in the same manner as we saw in the examples at the beginning of Section \ref{derive_patterns_section}.

The second pattern in Figure \ref{patterns_figure_2} (`$<$ \%7 10 $<$ \%3 $>$ $<$ \%4 $>$ $>$') is derived from the sequence `$<$ \%3 $>$ $<$ \%4 $>$' within the abstract pattern in row 2 of Figure \ref{alignment_3}. The third pattern (`$<$ \%8 11 $<$ \%1 $>$ $<$ \%2 $>$ $<$ \%7 $>$ $>$') is a new version of that abstract pattern that references the class of the second pattern (`\%7').

\subsubsection{Redundancy in Old}

Given that the system is dedicated to IC, it may seem strange that, at this stage of processing, there may be considerable replication of information (redundancy) amongst the patterns in Old. Patterns are added to Old but, at this stage, nothing is removed from Old. In the example just considered, the pattern `$<$ \%7 10 $<$ \%3 $>$ $<$ \%4 $>$ $>$' coexists with the pattern `$<$ \%5 7 $<$ \%1 $>$ $<$ \%2 $>$ $<$ \%3 $>$ $<$ \%4 $>$ $>$' even though they both contain the sequence `$<$ \%3 $>$ $<$ \%4 $>$'. In the example from Figures \ref{alignment_2} and \ref{patterns_figure_1}, the patterns `$<$ \%7 12 t h a t $>$', `$<$ \%9 14 b o y $>$' and `$<$ \%8 13 r u n s $>$ coexist with the pattern `$<$ \%1 9 t h a t b o y   r u n s $>$' despite the obvious duplication of information amongst these patterns.

The reason for designing the system in this way is that there is no guarantee that any given pattern derived from an alignment will ultimately turn out to be `correct' in terms of MLE principles or one's intuitions about what the correct grammar should be. Indeed, many of the patterns abstracted by the system are clearly `wrong' in these terms. Retention of older patterns in the store alongside patterns that have been derived from them leaves the door open for the system to create `correct' patterns at later stages regardless of whether `wrong' patterns had been created earlier. In effect, the system is able to explore alternative paths through the abstract space of possible patterns.

\subsection{Sifting and Sorting of Patterns}\label{sifting_and_sorting_section}

Identification of `wrong' patterns occurs in the {\em sifting\_and\_sorting()} stage of processing (operation 4 in Figure \ref{SP70_figure}), where the system develops one or more alternative grammars for the patterns in New in accordance with MLE principles. Figure \ref{sifting_and_sorting_figure} shows the overall structure of the {\em sifting\_and\_sorting()} function.

Each pattern in Old has an associated frequency of occurrence and, at the start of the function, all these values are set to zero. Then, all the patterns in New are reprocessed with the {\em create\_multiple\_alignments()} function, building multiple alignments as before, each one between one pattern from New and one or more patterns in Old. 

The difference on this occasion is that, for each CPFN, the best alignments are filtered to remove any that contain unmatched symbols in the CPFN or unmatched C symbols in any pattern from Old. The remaining `full' alignments provide the basis for further processing.

\begin{figure}[!hbt]
\fontsize{08.00pt}{09.60pt}
\begin{center}
\begin{BVerbatim}
SIFTING_AND_SORTING()
{
     1 For each pattern in Old, set its frequency of occurrence to 0.
     2 While (there are still unprocessed patterns in New)
     {
          2.1 Identify the first or next pattern from New as the CPFN.
          2.2 Apply the function CREATE_MULTIPLE_ALIGNMENTS() to
               create multiple alignments, each one between the CPFN
               and one or more patterns from Old.
          2.3 From amongst the best of the multiple alignments formed,
               select `full' alignments in which all the symbols of
               the CPFN are matched and all the C symbols are
               matched in each pattern from Old.
          2.4 For each pattern from Old, count the maximum number of
               times it appears in any one of the full alignments
               selected in operation 2.3. Add this count to the
               frequency of occurrence of the given pattern.
     }
     3 Compute frequencies of symbol types and their encoding costs.
          From these values, compute encoding costs of patterns in
          Old and new compression scores for each of the full
          alignments created in operation 2.
     4 Using the alignments created in 2 and the values computed in
          operation 3, COMPILE_ALTERNATIVE_GRAMMARS().
}
\end{BVerbatim}
\end{center}
\caption{The organisation of the {\em sifting\_and\_sorting()} function. The {\em compile\_alternative\_grammars()} function is described in Section \ref{compile_grammars}.}
\label{sifting_and_sorting_figure}
\end{figure}

In this phase of the program, we can be confident of finding at least one full alignment for each CPFN because, in the previous phase, each unmatched portion of the given pattern from New led to the creation of patterns in Old that would provide an appropriate match in the future.

When all the patterns from New have been processed in this way, there is a set $A$ of full alignments, divided into $b_1 ... b_m$ disjoint subsets, one for each PFN. From these alignments, the function computes the frequency of occurrence of each of the $p_1 ... p_n$ patterns in Old as:

\[f_i = \sum_{j = 1}^{j = m} max(p_i, b_j)\]

\noindent where $max(p_i, b_j)$ is the maximum number of times that $p_i$ appears in any {\em one} alignment in subset $b_j$. Using the maximum value for any {\em one} alignment for a given PFN is necessary because the alignments in each $b_j$ are {\em alternative} analyses of the corresponding PFN. If we simply counted the number of times each pattern appeared in all the alignments for a given PFN, the frequency values would be too high.

The function also compiles an alphabet of the symbol types, $s_1 ... s_r$, in the patterns in Old and, following the principles just described, computes the frequency of occurrence of each symbol type as:

\[F_i = \sum_{j = 1}^{j = m} max(s_i, b_j)\]

\noindent where $max(s_i, b_j)$ is the maximum number of times that $s_i$ appears in any {\em one} alignment in subset $b_j$.

From these values, the encoding cost of each symbol type is computed using the SFE method as before \citep{cover_thomas_1991}. As before (Section \ref{calculation_of_compression_scores}), the encoding cost of each of the `data' symbol types (those that appears in New) is weighted so that data symbols behave as if they were relatively large chunks of information.

Each symbol in each pattern in New and Old is then assigned the frequency and encoding cost of its type. With these values in place, the compression score of each alignment in the set of full alignments is recalculated.

Finally, in operation 4 of Figure \ref{sifting_and_sorting_figure}, a set of one or more alternative grammars is compiled, as described in Section \ref{compile_grammars}.

As the program stands, these alternative grammars are simply presented to the user for inspection. However, it is intended that the patterns in Old should be purged of all its patterns except those in the best grammar that has been found. It is anticipated that the program will be developed so that patterns from New will be processed in batches and that this kind of purging of Old will occur at the end of each batch to remove the `rubbish' and retain only those patterns that have proved useful in encoding the cumulative set of patterns from New.

\subsubsection{Compiling a Set of Alternative Grammars}\label{compile_grammars}

\sloppy The set of alternative grammars for the patterns in New are derived (in the {\em compile\_alternative\_grammars()} function) from the full alignments created in operation 2 of Figure \ref{sifting_and_sorting_figure}.

For any given PFN, a grammar for that pattern can be derived from one of its full alignments by simply listing the patterns from Old that appear in that alignment, counting multiple appearances of any pattern as one. Any such grammar may be augmented to cover an additional PFN by selecting {\em one} of the full alignments for the second PFN and adding patterns from Old that appear within that alignment and are not already present in the grammar (counting multiple appearances as one, as before). In this way, taking one PFN at a time, a grammar may be compiled for all the patterns from New.

A complication, of course, is that there are often two or more full alignments for any given PFN. This means that, for a given set of patterns from New, one can generate a tree of alternative grammars with branching occurring wherever there are two or more alternative alignments for a given PFN. Without some constraints, this tree can become unmanageably large.

In the {\em compile\_alternative\_grammars()} function, the tree of alternative grammars is pruned periodically to keep it within reasonable bounds. The tree is grown in successive stages, at each stage processing the alignments for {\em one} of the patterns from New and, for each grammar, processing only {\em one} alignment for each PFN. Values for $G$, $E$ and $T$ are calculated for each grammar and, at each stage, grammars with high values for $T$ are eliminated.

For a given grammar comprising patterns $p_1 ... p_g$, the value of $G$ is calculated as:

\[G = \sum_{i=1}^{i=g}(\sum_{j=1}^{j=L_i}s_j)\]

\noindent where $L_i$ is the number of symbols in the $i$th pattern and $s_j$ is the encoding cost of the $j$th symbol in that pattern.

Given that each grammar is derived from a set $a_1 ... a_n$ of alignments (one alignment for each PFN), the value of $E$ for the grammar is calculated as:

\[E = \sum_{i=1}^{i=n}e_i\]

\noindent where $e_i$ is the size, in bits, of the code string derived from the $i$th alignment (as described in Section \ref{NL_processing}), calculated as described in Section \ref{calculation_of_compression_scores}.

When the set of alternative grammars has been completed, each grammar is `cleaned up' by removing code symbols that have no function in the grammar and by renaming code symbols in a tidy manner (see Section \ref{example_1}, below).

Before leaving this section, it is worth pointing out a minor anomaly in the way values for $G$ and $E$ are calculated. These values depend on the encoding costs of symbols which themselves depend on frequencies of symbol types. At present, these frequencies are derived from the entire set of patterns in Old but it would probably be more appropriate if frequency values were derived from each grammar individually at each stage in the process of compiling grammars. The results are likely to be similar in both cases since encoding costs depend on relative frequency values, not absolute values and, for the symbol types appearing in any one grammar, it is likely that the ranking of frequency values derived from the grammar would be similar to the ranking derived from the entire set of patterns in Old.

\section{Evaluation of the Model}

Criteria that may be used to evaluate a learning model like SP70 include:

\begin{itemize}

\item If applications are scaled up to realistic size, is the model likely to make unreasonable demands for processing power or computer memory?

\item Given that the system aims to find grammars with relatively small values for $T$, does it succeed in this regard?

\item Does it produce knowledge structures that look `natural' or `reasonable'? Since humans are the most `powerful' learning systems on the planet, there is some justification for using human intuitions about the structuring of information as a touchstone for success or failure of an artificial learning system. Given evidence that human cognition is conditioned by MLE principles (Section \ref{MLE_and_intuition}), human intuitions provide an indirect check on whether or not MLE principles have been successfully implemented in the model.

\item Does it produce knowledge structures that successfully support other operations such as reasoning, making `expert' judgements, playing chess, and so on?

\end{itemize}

The first of these criteria is discussed in the next subsection. Evidence bearing on the second criterion is presented in Section \ref{plotting_values}, below, and the intuitive plausibility of results obtained from the model is considered at various points in Sections \ref{examples_section} and \ref{discussion}. No attempt has yet been made to evaluate SP70 in terms of the fourth criterion.

\subsection{Computational Complexity}\label{computational_complexity}

In common with other programs for unsupervised learning (and, indeed, other programs for finding good multiple alignments), SP70 does not attempt to find theoretically ideal solutions. This is because the abstract space of possible grammars (and the abstract space of possible alignments) is, normally, too large to be searched exhaustively. In general, heuristic techniques like hill climbing, genetic algorithms, simulated annealing etc must be used. By using these techniques, one can normally convert an intractable computation into one with computational complexity that is within acceptable limits.

\sloppy In SP70, the critical operation is the formation of multiple alignments ({\em create\_multiple\_alignments()}). Other operations (e.g., the {\em derive\_patterns()} function) are, in comparison, quite trivial in their computational demands.

\sloppy In a serial processing environment, the time complexity of the {\em create\_multiple\_alignments()} function has been estimated \citep{prob_reason_report} to be approximately O$(log_2 n \times nm)$, where $n$ is the size of the pattern from New (in bits) and $m$ is the sum of the lengths of the patterns in Old (in bits). In a parallel processing environment, the time complexity may approach O$(log_2 n \times n)$, depending on how well the parallel processing is applied. In serial and parallel environments, the space complexity should be
O$(m)$.

In SP70, the function is applied (twice) to the set of patterns in New so we need to take account of how many patterns there are in New. It seems reasonable to assume that the sizes of patterns in New are approximately constant.\footnote{There is no requirement in the model that patterns in New should, for example, be complete sentences. They may equally well be arbitrary portions of incoming data, perhaps measured off by some kind of input buffer.}

Old is initially empty and grows as learning proceeds. The size of Old (before purging) is, approximately, a linear function of the size of New. Given this growth in the size of Old, the time required to create alignments for any given pattern from New will grow as learning proceeds. Again, the relationship is approximately linear. 

So if we ignore operations other than the {\em create\_multiple\_alignments()} function, we may estimate the time complexity of the program (in a serial environment) to be O$(N^2)$ where $N$ is the number of patterns in New. In a parallel processing environment, the time complexity may approach O$(N)$, depending on how well the parallel processing is applied. In serial or parallel environments, the space complexity should be O$(N)$.

The time complexity of the program may be improved when it has been developed, as envisaged, so that the New patterns are processed in batches, with a purging of Old between each batch to remove all patterns except those in the best grammar. In this case, the size of each batch of New patterns will be approximately constant and the running time of the program should be roughly proportional to the number of batches of New patterns.

\section{Examples}\label{examples_section}

This section presents two examples to illustrate how SP70 works and what it can do. The examples are fairly simple, partly for the sake of clarity and partly because of shortcomings in the model (discussed in Section \ref{discussion}, below).

\subsection{Example 1}\label{example_1}

The four short sentences supplied to SP70 as New for this first example are shown in Figure \ref{example_1_patterns}.

\begin{figure}[!hbt]
\begin{center}
\begin{BVerbatim}
j o h n r u n s
m a r y r u n s
j o h n w a l k s
m a r y w a l k s
\end{BVerbatim}
\end{center}
\caption{Four patterns supplied to SP70 as New.}
\label{example_1_patterns}
\end{figure}

The two best grammars found by the program for these sentences (with 10 as the cost factor) are shown in Figure \ref{example_1_grammars}. In (a) the best grammar is shown in the form that it is first compiled and in (b) the same grammar is shown after it has been cleaned up. Figure \ref{example_1_grammars} (c) shows the second-best grammar after it has been cleaned up.

\begin{figure}[!hbt]
\begin{center}
\begin{BVerbatim}
< %4 15 s >
< %7 %9 %11 152 m a r y >
< %9 %14 162 j o h n >
< %24 406 r u n >
< %24 %27 407 w a l k >
< %25 412 < %9 > < %24 > < %4 > >

(a)

< %2 1 s >
< %3 2 m a r y >
< %3 3 j o h n >
< %1 4 r u n >
< %1 5 w a l k >
< 6 < %3 > < %1 > < %2 > >

(b)

< %2 2 m a r y >
< %2 3 j o h n >
< %1 1 r u n s >
< %1 5 w a l k s >
< 4 < %2 > < %1 > >

(c)
\end{BVerbatim}
\end{center}
\caption{Grammars found by SP70 with the four patterns shown in Figure \ref{example_1_patterns} supplied as New. (a) The best grammar without cleaning up. (b) The same grammar after cleaning up. (c) The second best grammar after cleaning up.}
\label{example_1_grammars}
\end{figure}

Cleaning up grammars means removing class symbols that have no referents (e.g., `\%7' and `\%11' in the second pattern in Figure \ref{example_1_grammars} (a)) and renumbering the class symbols and the discrimination symbols, starting from 1 for each set. The renumbering is a purely cosmetic matter and makes no difference to the encoding cost calculated for each symbol.

The two grammars shown in Figure \ref{example_1_grammars} are both reasonably plausible grammars for the four original sentences. In the best grammar `m a r y' and `j o h n' are picked out as discrete words and assigned to the same grammatical class (`\%3'). In a similar way, `r u n' and `w a l k' are each picked out as a discrete entity---corresponding to the `stem' of a verb---and both are assigned to the class `\%1'. The suffix for these verb stems (`s') is picked out as a distinct entity and the overall sentence structure is captured in the pattern `$<$ 6 $<$ \%3 $>$ $<$ \%1 $>$ $<$ \%2 $>$ $>$'. The second best grammar is the same except that the suffix of each verb is not separated from the stem.

\subsubsection{Intermediate Results}

The simplicity of the results shown in Figure \ref{example_1_grammars} disguises what the program has done to achieve them. The flavour of this processing may be seen from the selection of intermediate results presented here.

When the first pattern from New is processed, Old is empty except for a copy of that first pattern that is added to Old, one symbol at a time, as the pattern is processed. The only alignment formed at this stage is shown in Figure \ref{alignment_4}. Remember that the symbols in the PFN (`j o h n r u n s') must not be aligned with the corresponding symbols in the pattern in Old because, in effect, this would mean matching each symbol with itself (Section \ref{copying_CPFN}).

\begin{figure}[!hbt]
\begin{center}
\begin{BVerbatim}
0 j o h n r u  n s         0
               |          
1 < %1 5 j o h n r u n s > 1
\end{BVerbatim}
\end{center}
\caption{The only alignment formed when the first pattern from New is processed.}
\label{alignment_4}
\end{figure}

It is evident that, at this stage, opportunities to gain any useful insights into the overall structure of the patterns in New are quite limited. From the `bad' alignment shown in Figure \ref{alignment_4} the program abstracts the `bad' patterns `$<$ \%2 7 n $>$', `$<$ \%3 10 j o h $>$', `$<$ \%3 11 j o h n r u $>$', `$<$ \%4 14 r u n s $>$', `$<$ \%4 15 s $>$' and `$<$ \%5 19 $<$ \%3 $>$ $<$ \%2 $>$ $<$ \%4 $>$ $>$'.\footnote{As applied to alignments and patterns, the words `bad' and `good' are shorthand for ``bad/good in terms of MLE principles and perhaps also in terms of one's intuitions about what is or is not an appropriate grammar for the data''. Quote marks will be dropped in the remainder of the paper.}

When the next PFN (`m a r y r u n s') is processed, the program is able to form more sensible alignments like

\begin{center}
\begin{BVerbatim}
0 m a r y r u n s   0
          | | | |  
1 < %4 14 r u n s > 1
\end{BVerbatim}
\end{center}

\noindent and

\begin{center}
\begin{BVerbatim}
0 m a r y        r u n s   0
                 | | | |  
1 < %1 5 j o h n r u n s > 1.
\end{BVerbatim}
\end{center}

The first of these alignments yields the patterns `$<$ \%7 152 m a r y $>$' and `$<$ \%8 157 $<$ \%7 $>$ $<$ \%4 $>$ $>$'. It would have created a pattern for `r u n s' but this is suppressed because the program detects that a pattern with those C symbols already exists (`$<$ \%4 14 r u n s $>$').

From the second alignment, the system derives the pattern `$<$ \%9 162 j o h n $>$' (assigned to the class `\%9') and it would create a pattern for `m a r y' but it detects that a pattern with these C symbols already exists (`$<$ \%7 152 m a r y $>$'). However, since `j o h n' and `m a r y' occur in the same context (`---r u n s'), they should be assigned to the same context-defined class. Accordingly, the program adds the class symbol `\%9' to the pattern `$<$ \%7 152 m a r y $>$' so that it becomes `$<$ \%7 \%9 152 m a r y $>$' and it creates the abstract pattern `$<$ \%10 168 $<$ \%9 $>$ $<$ \%4 $>$ $>$', tying the whole structure together.

As processing proceeds in the pattern-generation phase (operation 3 in Figure \ref{SP70_figure}), the program forms good alignments like

\begin{center}
\begin{BVerbatim}
0        j o h n w a l k s   0
         | | | |         |  
1 < %1 5 j o h n r u n   s > 1
\end{BVerbatim}
\end{center}

\noindent and relatively bad ones like

\begin{center}
\begin{BVerbatim}
0        j o h       n w a l k s   0
         | | |       |         |  
1 < %1 5 j o h n r u n         s > 1.
\end{BVerbatim}
\end{center}

\noindent From alignments like these it derives correspondingly good and bad patterns.

By the time the last PFN is processed (`m a r y w a l k s') there are enough good patterns in Old for the program to start forming quite plausible alignments like

\begin{center}
\fontsize{08.00pt}{09.60pt}
\begin{BVerbatim}
0                          m a r y                        w a l k s     0
                           | | | |                        | | | | |    
1                          | | | |   < %22 %29 %4 %31 394 w a l k s >   1
                           | | | |   |         |                    |  
2 < %8 157 < %7            | | | | > <         %4                   > > 2
           | |             | | | | |                                   
3          < %7 %9 %11 152 m a r y >                                    3.
\end{BVerbatim}
\end{center}

At the end of this phase of processing, Old contains a variety of patterns including several good ones and quite a lot of bad ones.

In the second {\em sifting\_and\_sorting()} phase of the program (operation 4 in Figure \ref{SP70_figure}) the program compiles a set of alternative grammars the best of which are shown in Figure \ref{example_1_grammars}.

\subsubsection{Plotting Values for $G$, $E$ and $T$}\label{plotting_values}

The value of $T$ for the best grammar is 1379 bits before cleaning up and 1348 bits after cleaning up. For the second-best grammar, the corresponding values are 1418 and 1377. By contrast, $T$ is calculated to be 2245 bits for the `naive' grammar that comprises the four patterns from New with added ID symbols (`$<$ \%1 5 j o h n r u n s $>$', `$<$ \%6 21 m a r y r u n s $>$', `$<$ \%21 212 j o h n w a l k s $>$' and `$<$ \%33 457 m a r y w a l k s $>$').

\begin{figure}[!hbt]
\centering
\includegraphics[width=1\textwidth,height=0.5\textheight]{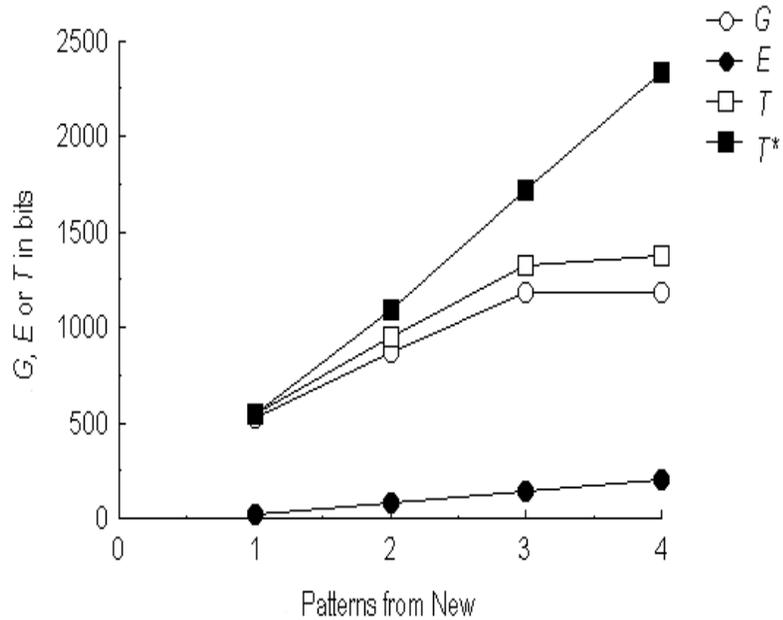}
\caption{Values of $G$, $E$ and $T$ for the best grammar found (before cleaning up) as successive patterns from New are processed in {\em compile\_alternative\_grammars()}. Values for $T$ for the `naive' grammar are also shown (marked with `*').}
\label{plotting_figure}
\end{figure}

Figure \ref{plotting_figure} shows how the values of $G$, $E$ and $T$ change as successive patterns from New are processed when the set of alternative grammars is compiled. Each point on each of the lower three graphs represents the relevant value from the best grammar found after the full alignments for a given PFN have been processed. The top graph shows successive values of $T$ for the `naive' grammar mentioned above.

Notice how the value of $G$ does not change between the third PFN and the fourth. This is because there is nothing in the fourth pattern that has not already been found in the first three patterns.

\subsection{Example 2}\label{example_2}

When New contains the eight sentences shown in Figure \ref{example_2_patterns}, the best grammar found by SP70 (after cleaning up) is the one shown in Figure \ref{example_2_grammar}.

\begin{figure}[!hbt]
\begin{center}
\begin{BVerbatim}
t h a t b o y r u n s
t h a t g i r l r u n s
t h a t b o y w a l k s
t h a t g i r l w a l k s
s o m e b o y r u n s
s o m e g i r l r u n s
s o m e b o y w a l k s
s o m e g i r l w a l k s
\end{BVerbatim}
\end{center}
\caption{Eight sentences supplied to SP70 as New.}
\label{example_2_patterns}
\end{figure}

\begin{figure}[!hbt]
\begin{center}
\begin{BVerbatim}
< %2 2 s o m e >
< %2 3 t h a t >
< %1 5 b o y >
< %1 6 g i r l >
< %3 4 r u n s >
< %3 7 w a l k s >
< 1 < %2 > < %1 > < %3 > >
\end{BVerbatim}
\end{center}
\caption{The best grammar found by SP70 (after cleaning up) when New contains the eight sentences shown in Figure \ref{example_2_patterns}.}
\label{example_2_grammar}
\end{figure}

This result looks reasonable but, in the light of the best grammar found in Example 1, one may wonder why the terminal `s' of `r u n s' and `w a l k s' has not been identified as a discrete entity, separate from the verb stems `r u n' and `w a l k'.

In the pattern-generation phase of processing, SP70 does form alignments like this

\begin{center}
\begin{BVerbatim}
0        t h a t b o y w a l k s   0
         | | | | | | |         |  
1 < %1 9 t h a t b o y r u n   s > 1
\end{BVerbatim}
\end{center}

\noindent which clearly recognises the verb stems and `s' as distinct entities. But for reasons that are still not entirely clear, the program does not build these entities into plausible versions of the full sentence structure. The program isolates the pattern `$<$ \%25 599 t h a t b o y $>$' from the alignment just shown but for some reason it fails to find internal structure within that pattern---although it recognises `t h a t' and `b o y' in other contexts. This issue is discussed in Section \ref{finding_internal_structure}, below.

\section{Discussion}\label{discussion}

This section discusses a selection of issues relating to the SP70 model, especially shortcomings of the current version and how they may be overcome.

\subsection{Finding Internal Structure}\label{finding_internal_structure}

The failure of the model to find some of the internal structure within a sentence in Example 2 seems to be a manifestation of a more general shortcoming. Although the model in its current form can isolate basic segments and tie them together in an overall abstract structure, it is not good at finding intermediate levels of abstraction.

What seems to be needed is some kind of additional reprocessing of the patterns in Old, including the abstract patterns that have been added to Old, to discover partial matches not detected in the initial processing of the patterns from New. This should allow the system to detect intermediate levels of structure such as phrases or clauses or structures that may exist within smaller units such as words.

\subsection{Finding Discontinuous Dependencies}

In the development of the model to date, no attempt has been made to enable the system to detect discontinuous dependencies such as number dependency between the subject of a sentence and its main verb (as mentioned in Section \ref{NL_processing} and illustrated in Figure \ref{alignment_1}) or gender dependencies in languages like French.

Although this kind of capability may seem like a refinement that we can afford to do without at this stage of development, a deficiency in this area seems to have an impact on the program's performance at an elementary level. Even in quite simple structures, dependencies can exist that bridge intervening structure and, in its current form, the program does not encode this kind of information in a satisfactory manner.

There do not seem to be any insuperable obstacles to solving this problem within the ICMAUS framework:

\begin{itemize}

\item The format for representing knowledge accommodates these kinds of structure quite naturally.

\item Finding partial matches that bridge intervening structure is bread-and-butter for the {\em create\_multiple\_alignments()} function.

\end{itemize}

\noindent What seems to be required is some revision of the way in which patterns are derived from alignments.

\subsection{Generalization of Grammatical Rules and the Correction of Overgeneralizations}\label{generalization}

A well-documented phenomenon in the way young children learn language is that they say things like ``The ball was hitted'' or ``Look at the gooses'', apparently applying general rules for constructing words but applying them too generally. How is it that children eventually learn to avoid these kinds of overgeneralizations? It is tempting to suppose that children are corrected by parents or other adults but the weight of empirical evidence is that, while such corrections may be helpful, they are not actually necessary for language learning.

MLE principles provide an elegant solution to this puzzle. Without the kind of feedback or supplementary information postulated by \citet{gold_1967}, it is possible to search for grammars that are good in MLE terms and these are normally ones that steer a path between generalizations that are, intuitively, `correct' and others that appear to be `wrong'. This kind of effect has been demonstrated with the SNPR model \citep{wolff_1982}.

When SP70 is run on the first three of the four patterns shown in Figure \ref{example_1_patterns}, the best grammar found is exactly the same as before (Figure \ref{example_1_grammars} (b)). This grammar generates the missing sentence (`m a r y w a l k s') as well as the other three sentences, but it does not generate anything else.

In this example, the model generalizes in a way that seems intuitively to be correct and avoids creating overgeneralizations that are glaringly wrong. However, relatively little attention has so far been given to this aspect of the model and further work is required. In particular, a better understanding is needed of alternative ways in which grammatical rules may be generalized.

\subsection{Other Developments}

Other areas where further work is planned include:

\begin{itemize}

\item As was indicated in Section \ref{sifting_and_sorting_section}, it is anticipated that the program will be developed so that it processes patterns from New in batches, purging bad patterns from Old at the end of each batch.

\item As was noted in Section \ref{avoiding_duplication}, it is possible for two or more context-defined classes to be the same. The program needs to check for this possibility and merge identical classes whenever they are found.

\item At present, the program applies the {\em create\_multiple\_alignments()} function twice to each PFN, once as part of the process of generating patterns to be added to Old and once in the {\em sifting\_and\_sorting()} phase. It seems possible that the two phases could be integrated so that the {\em create\_multiple\_alignments()} function need only be applied once to each PFN.

\item As was suggested in Section \ref{derive_patterns_section}, there may be a case for introducing a `null' pattern to allow for the encoding of optional elements in a syntactic structure.

\item Although the computational complexity of the model on a serial machine is within acceptable limits, improvements in that area, and higher absolute speeds, may be obtained by the application of parallel processing. If or when residual problems in the model have been solved, it is envisaged that the system will be developed as a software virtual machine on existing high-parallel hardware or perhaps on new forms of hardware dedicated to the needs of the model. It may be possible to exploit optical techniques to achieve high-parallel matching of patterns in the core of the model.

\end{itemize}

\subsection{Motivation and emotion}

Although learning has been considered in this article primarily as an engineering problem, the theory that has been described may be viewed as a possible theory of learning in people and other animals.

We tend to remember things best that are significant for us in terms of our motivations and emotions. How would this fit in with the kinds of learning mechanisms that have been described?

The tentative suggestion here is that motivations and emotions may have an impact when patterns are purged from the system. Other things being equal, we may suppose that MLE principles govern the choice of which patterns should be retained and which should be discarded. But any pattern that represents something that has special significance may be retained by the system even if it does not score well in terms of MLE measures.

\section{Conclusion}

Although SP70 is still some way short of an `industrial strength' system for unsupervised learning, the results obtained so far are good enough to show that the general approach is sound. Problems that have been identified appear to be soluble.

A particular attraction of this approach to learning is that the ICMAUS framework provides a unified view of a variety of issues in AI thus facilitating the integration of learning with other aspects of intelligence.

\section*{Acknowledgements}

I am very grateful for constructive comments that I have received from Pat Langley.

\end{document}